\documentclass{article}
\usepackage{spconf,amsmath,graphicx,booktabs,verbatim}
\usepackage[T1]{fontenc}
\usepackage{enumitem}
\usepackage{algorithm}
\usepackage{algorithmic}
\usepackage{svg}
\usepackage{arydshln} 
\setenumerate[1]{itemsep=0pt,partopsep=0pt,parsep=\parskip,topsep=5pt}
\setitemize[1]{itemsep=0pt,partopsep=0pt,parsep=\parskip,topsep=5pt}
\setdescription{itemsep=0pt,partopsep=0pt,parsep=\parskip,topsep=5pt}

\title{FedVMR: A New Federated Learning method for Video Moment Retrieval}
\name{Yan Wang$^{1}$, Xin Luo$^{1}$, Zhen-Duo Chen$^{1}$, Peng-Fei Zhang$^{2}$, Meng Liu$^{3}$, Xin-Shun Xu$^{1}$}
\address{$^{1}$School of Software, Shandong University  \\ $^{2}$School of Information Technology and Electrical Engineering, University of Queensland \\ $^{3}$School of Computer Science, Shandong Jianzhu University}

\begin{document}
%
\maketitle
\begin{abstract}
Despite the great success achieved, existing video moment retrieval (VMR) methods are developed under the assumption that data are centralizedly stored. However, in real-world applications, due to the inherent nature of data generation and privacy concerns, data are often distributed on different silos, bringing huge challenges to effective large-scale training. In this work, we try to overcome above limitation by leveraging the recent success of federated learning. As the first that is explored in VMR field, the new task is defined as video moment retrieval with distributed data. Then, a novel federated learning method named FedVMR is proposed to facilitate large-scale and secure training of VMR models in decentralized environment. Experiments on benchmark datasets demonstrate its effectiveness. This work is the very first attempt to enable safe and efficient VMR training in decentralized scene, which is hoped to pave the way for further study in the related research field. 

\end{abstract}
\section{Introduction}
\label{sec:intro}
Video moment retrieval (VMR)~\cite{2d-tan,mcn,structured,fastvmr,rwm,avmr,umt}, aiming to locate the target video moment that best corresponds to the given query sentence of natural language from a long video, has become one of the most intriguing and hot topics in video understanding literature. While promising results are achieved, almost all existing VMR methods are designed for centralized data~\cite{vmrsurvey,vmrsurvey2}, deterring them from real-world applications. Videos in the reality are often created and stored with personal cameras, CCTV systems or other distributed devices. In such scenarios, aggregating data from different devices or datasets to enable large-scale training of VMR will be faced with the challenge of an expensive cost of transmission and storage. Furthermore, as videos might also contain sensitive information, sharing them would inevitably cause information leakage, leading to a serious data privacy problem. 

Federated learning has recently achieved much attention, which can provide powerful and safe conditions for large-scale model training \cite{fedavg,icasspfederated,scaffold}. The purpose of federated learning is to train a model with decentralized data on various clients by aggregating the local updates of model parameters instead of direct data collection. Considering this, we novelly introduce federated learning into the VMR domain and define a new research problem termed \textbf{video moment retrieval with distributed data}. The problem here is the non-negligible harm to the performance and efficiency of the resulting models from distributed learning. Traditional federated learning methods cannot well preserve knowledge learned because each client trains model in parallel without interaction, which limits the performance improvement and convergence speed during training.

Considering these, this paper proposes a new method dubbed \textbf{Fed}erated \textbf{V}ideo \textbf{M}oment \textbf{R}etrieval (abbr. \textbf{FedVMR}), to enable high-efficiency and effective VMR modeling training with distributed data. Specifically, to cope with the performance degradation caused by distributed learning, a new strategy called \textbf{grouped sequential federated learning} is proposed. It divides clients into different groups, where clients in each group are trained sequentially by taking advantage of learning experiences from peers. In the meantime, a novel selective model aggregation method is designed to enhance the effect of effective local models while penalizing inferior local models. Considering the non-IID problem, a temporal-distribution-gap loss is devised to bridge distribution gaps among different clients. The main contributions of this paper are as follows: 
\begin{itemize}[leftmargin=*]
\item \textbf{Task contribution.} To the best of our knowledge, we are the first to propose a novel task, i.e., video moment retrieval with federated learning, to facilitate large-scale and secure VMR training with distributed data.


\item \textbf{Technical contribution.} A novel method termed FedVMR is proposed to solve the new task to deal with the performance degradation in distribution learning and the distribution gaps among different clients. Extensive experiments on two benchmark datasets demonstrate the superiority of the proposed FedVMR. 

\item \textbf{Community contribution.} As far as we know, the grouped sequential federated learning is the first attempt in federated learning community to design methods in a sequential manner. We hope this work would facilitate future studies. 

\end{itemize}

\section{Methodology}
\label{sec:model}


\subsection{Notations and Problem Definition}

 Given a video and a query sentence, the target of video moment retrieval models is to locate the video moment that the query sentence best describes. More specifically, the corresponding temporal indexes of the moment, i.e., the start index and the end index, should be returned. To learn those models, training data, i.e., videos which contain multiple ground truth annotations with query and corresponding time indexes, are needed. Traditional VMR task implicitly assumes those data could be collected. In other words, a centralized dataset $D$ is constructed and used to train the model $w_{centralized}$.

In real applications, videos are usually stored in different places. Thus, the task learned in this paper, {i.e., video moment retrieval with distributed data}, aims to train VMR models with all client's data without privacy disclosure. In particular, suppose there are $C$ clients, which have different corpus of videos $\{D_{1},...,D_{C}\}$. Let $w_{i}$ and $w_{global}$ denote the model parameters of the $i$-th $(i=\{1,\cdots,C\})$ client and global server, respectively. The aim is to learn the VMR model parameterized by $w_{global}$ which is aggregated from  $w_{i}$ $(i=\{1,\cdots,C\})$ by the trusted central server. Ideally, the learnt model could perform as well as that trained on the centralized dataset $D$.


\begin{figure}
	\centering
	\includegraphics[width=0.42\textwidth]{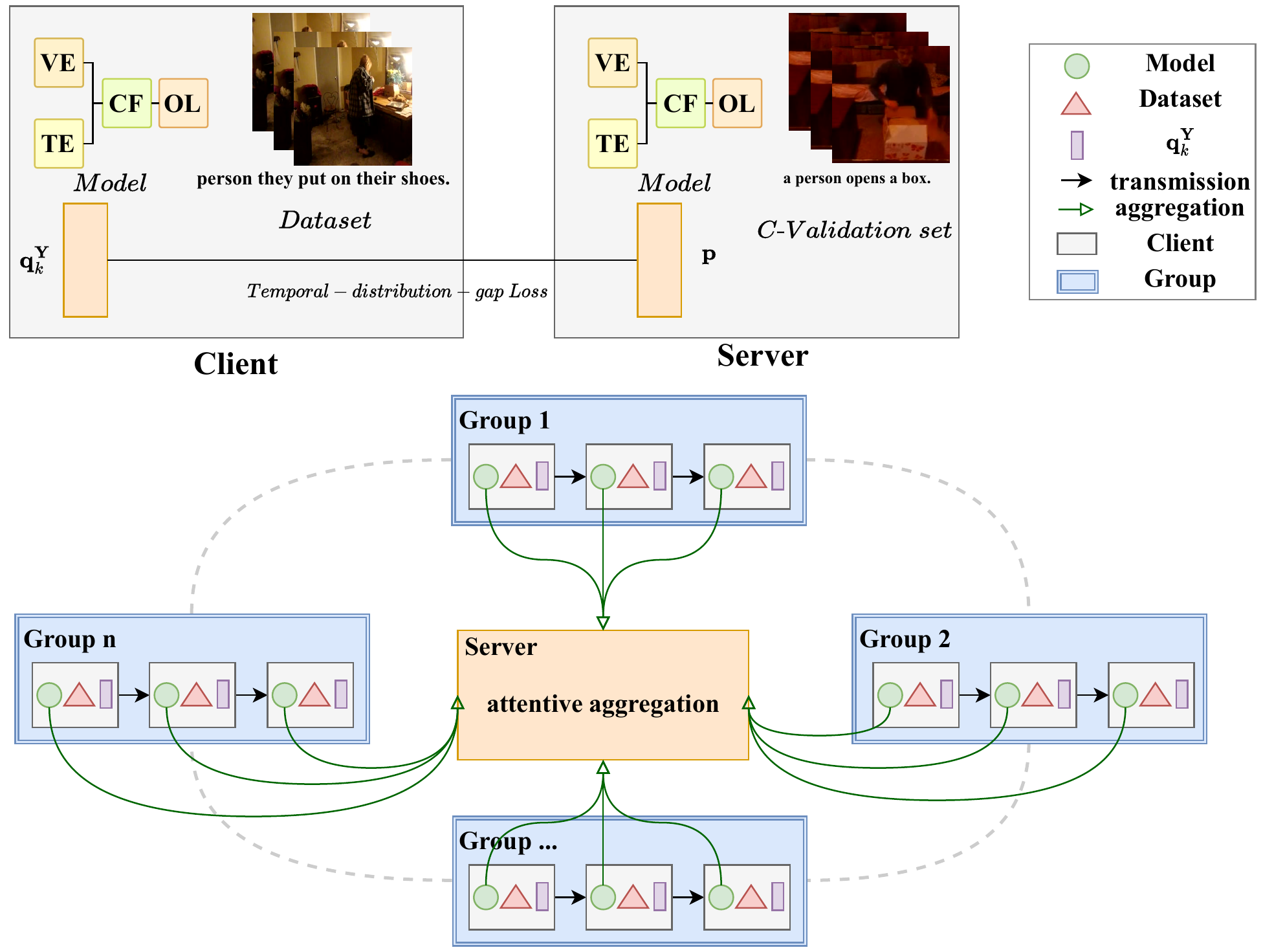}
	\caption{ An illustration of the proposed FedVMR. The proposed method divides clients into different groups, where sequential learning is performed within each group while parallel training is conducted among groups.
}
	\label{model}
\end{figure}




\begin{algorithm}[tb]
	\caption{FedVMR.}
	\label{alg:algorithm}
	\textbf{Input}: number of clients $C$, number of communication rounds $R$, number of groups $G$.\\
	\textbf{Output}: model parameters $w^{R}_{global}$ after $R$ rounds.

	\begin{algorithmic}[0] 
		\STATE  \textbf{Server executes:}
		\STATE  initialize $w^{0}$  
		\FOR{$k=1,2, \ldots, C$}
		\STATE  Calculate the temporal class distribution $\textbf q_{k}$ of client $k$ using the temporal annotations of client $k$'s own dataset
		\ENDFOR
		\STATE  $\textbf p \leftarrow\sum_{k} \frac{n_{k}}{n} \textbf q_{k}$.
		
		\STATE Randomly divide all clients to $G$ groups 
		\STATE Construct c-validation set

		\FOR{ round $t=1,2, \ldots, R$}
		\STATE  Send global model $w^{t}_{global}$ to clients
		\FOR{each group \textbf{in parallel}}
		\FOR{each client in current group \textbf{in turn}}
		\STATE  $w_{k}^{t}$ $\leftarrow$  \textbf{ClientUpdate} (\textit{current clinet})
		\STATE  $a_k^{t} \leftarrow$ verify current model on {\it c-validation set}
		\ENDFOR
		\ENDFOR
		
		\STATE  $w^{t+1}_{global} \leftarrow \sum_{k=1}^{C} a_{k}^{t} w_{k}^{t}.$
		\ENDFOR
		\STATE  \textbf{return} $w^{R}_{global}$
		\STATE  \textbf{ClientUpdate:  } \textit{\% Train VMR model in a local client}.
		\STATE  Initialize with model from last client of the same group
		\STATE  Calculate the predicted temporal class distribution $\textbf{q}_{k}^\textbf{Y}$
		\STATE  $L_{dis}$ $\leftarrow$ KL divergence of  $\left(\textbf p, \textbf q_{k}^Y\right)$
		\STATE  Update model parameters
		\STATE  Transmit model to server and next client of the same group
		\STATE  \textbf{return}
	\end{algorithmic}
\end{algorithm}

\subsection{FedVMR}
To enable effective training in the decentralized environment, a new method called FedVMR is proposed, where an illustration of the framework can be found in Figure \ref{model}. As the work focuses on solving problems by designing a new learning scheme instead of new models, we use visual encoder (VE), text encoder (TE), a cross-modal fusion module (CF), and an output layer (OL) to represent the general VMR models in this figure. Similar to the standard setting of federated learning, there exists one trusted server and $C$ clients. FedVMR first divides $C$ clients into $G$ groups. Then, grouped sequential federated learning is conducted, where sequentialization is performed within each group and parallelization is conducted among groups. To better merge knowledge from different clients, the proposed method quantificationally measures the quality of local models by constructing a tiny dataset, and accordingly penalizes inferior local models and pays more attention to better local models. At the same time, a new strategy called temporal-distribution-gap loss is designed to solve the key issue of federated learning, i.e., the non-IID problem. The overall algorithm of our FedVMR is shown in Algorithm~\ref{alg:algorithm}. Details are presented as follows.

\subsection{Grouped Sequential Federated Learning} 

Traditional federated learning, which usually trains all local models in parallel and aggregates resulting models, often suffers from performance deterioration due to the lack of interaction between different clients. To ensure the final performance while preserving efficiency and security, FedVMR introduces a new strategy named grouped sequential federated learning. More specifically, before training, all $C$ clients are divided into $G$ groups, and the training of models between different groups is conducted in parallel. Within each group, models are trained sequentially, where the model parameters of the last client are transmitted to the next client as the initialization parameters. By taking advantage of knowledge from other clients, it is expected to obtain a high-performance model. Compare to traditional federated learning, the proposed method can effectively improve the performance of target models, while preserving security and efficiency.  

\subsection{C-Validation Set} 

With well-trained local models, the server would aggregate them together to gain a global model and then use them to update local models until desired results. Conventional federated learning gains the global model by simply averaging local models without differentiating the model quality, which may lead to unsatisfactory performance. To solve this, our solution is to enhance the effect of better local models while weakening the other models. To this end, we propose to measure the model performance trained on different local clients by introducing a new strategy called {c-validation set}.



C-validation set is composed of a tiny fraction of the training data voluntarily uploaded by clients, used to validate the performance of decentralized models trained on clients. In FedVMR, we define the validation function of the $k$-th client as the weighted average of IoU indicators $a_k =F(\sum_{h} IoU_h * e_h)$,
where $IoU_h$ indicates the percentage of the results having IoU larger than $h$, $e_h$ is the weight assigned to each IoU indicator, and $F$ represents the softmax function. By substituting the IoU value into $a_k$, we can get the attention score of client $k$. Then, the aggregation of FedVMR is $w^t_{global} = \sum_{k=1}^C {a_k^t} * {w_k^t}$, where superscript $t$ represents the round $t$.

\subsection{Temporal-Distribution-Gap Loss} 

Another issue in federated learning is the non-IID problem, deriving from the difference in label distributions among different clients. Traditional federated learning is usually built upon classification tasks, and the extent of the non-IIDness can be easily described with the availability of explicit labels. However, in VMR tasks, temporal indexes of moments are more concerned with than video-level labels, making it hard to measure the non-IIDness~\cite{noniid,noniid2}. To tackle the challenge, we construct the temporal class distribution.


VMR methods focus on the start index and end index of the video moment. According to the location of timepoints, we assign a temporal class to each video. For example, we can define two temporal classes for the start point by considering the location this timepoint falls into, i.e., the first half or the second half of this video. The same goes for the end timepoint. By taking the location of both the start and end timepoint into consideration, we can divide the whole dataset into $4$ temporal classes. Then, the temporal class distribution $\textbf q_k$ of the client $k$ can be defined as $\textbf q_k = \left \{ p_1, p_2, ..., p_i, ... \right \} $, where $p_i$ represents the probability of temporal class $i$.



Specifically, we take the temporal class distribution $\textbf q_k^\textbf{Y}$ according to the prediction $\textbf{Y}$ of client $k$ as a measure of the gap among clients. Inspired by \cite{2019arxiv}, we define the population’s class distribution as $	\textbf p = \sum_{k} \frac{n_{k}}{n}  \textbf q_{k}$. Then we apply KL divergence to construct the loss function of $\textbf q_k^\textbf{Y}$ and $\textbf p$, participating in the model update of client $k$, $L_{dis}^k = - \sum_{x \in \chi} \textbf q_k^\textbf{Y}(x) \log {\frac{\textbf q_k^\textbf{Y}(x)}{\textbf p(x)}}$,
where $\chi$ represents the set of all temporal classes and $x$ represents a certain class of $\chi$. In real situations, $\textbf q_k(x)$ or $\textbf p(x)$ may be 0, which violates the mathematical principles. Therefore, we set $	\textbf q^k =\textbf q^k + \textbf{1}, $
where $\textbf{1}$ is a vector of ones.  


\section{Experiments}
\label{sec:exp}

\subsection{Experimental Settings}

\textbf{Datasets.} Two commonly-used benchmark datasets in the VMR domain are adopted.  \textbf{Charades-STA}~\cite{ctrl} is constructed on Charades~\cite{charades} dataset, which consists of $9,848$ videos of indoor activity. It contains $12,408$ pieces of data in the training set and $3,720$ in the testing set. \textbf{ActivityNet Captions}~\cite{activitynet} contains $19,209$ videos. Following~\cite{cmin}, we split this dataset as $37,421$, $17,505$, and $17,031$ moment-sentence pairs for training, validation, and testing. 


\noindent\textbf{Metrics.} Following the setting in the video moment retrieval task, we chose $R(1, m)$ as an indicator of performance, which means the percentage of the results having Intersection over Union (IoU) larger than $m$ ($m \in (0, 1]$).
 
\noindent\textbf{Baselines.}
Four methods are selected as our baselines, including three state-of-the-art federated learning methods, FedAvg~\cite{fedavg}, FedProx~\cite{fedprox}, and FedCMR~\cite{fedcmr}, and a centralized method. The centralized VMR method is to train the model on the centralized dataset, i.e., collecting data from all clients into a dataset. Its performance acts the gold standard.


\begin{table*}
	 \scriptsize
	\centering
	
		\begin{tabular}{lccccc|lccccc}
			\toprule[1pt]
			\multicolumn{6}{c|}{Charades-STA} & \multicolumn{6}{c}{ActivityNet Captions}\\
			\midrule
			Metric & Centralized & FedAvg & FedProx & FedCMR & FedVMR & Metric & Centralized & FedAvg & FedProx & FedCMR & FedVMR\\
			\midrule
			IoU$>$0.7 & 30.15 & 25.54 & 28.15 & 19.72 & \textbf{31.60} & IoU$>$0.7 & 22.74 & 17.56 & 17.52 & 4.19 & \textbf{22.73}\\
			IoU$>$0.5 & 53.29 & 50.51 & 51.37 & 39.41 & \textbf{53.01} & IoU$>$0.5 & 42.49 & 36.88 & 36.81 & 20.01 & \textbf{42.35}\\
			IoU$>$0.3 & 67.19 & 66.00 & 65.22 & 55.98 & \textbf{66.46} & IoU$>$0.3 & 61.41 & 53.52 & 53.32 & 45.59 & \textbf{61.67}\\
			\toprule[1pt]
	\end{tabular}
	\caption{The accuracy of all methods on Charades-STA and ActivityNet Captions in the case of $G=1$. }
	\label{tab:baseline}
\end{table*}
 
\noindent\textbf{VMR Model.} As this paper focuses on training VMR models with decentralized data rather than designing a specific VMR model, we leveraged the existing VMR method SSMN~\cite{ssmn} in experiments.



\noindent\textbf{Implementation Details.} $e_h$ is set to $0.1$ for the indicator of IoU$>0.1$, $0.2$ for IoU$>0.3$, $0.2$ for IoU$>0.5$, $0.4$ for IoU$>0.7$, and $0.1$ for IoU$>0.9$. To construct the voluntarily uploaded c-validation set, we randomly chose $1\%$ videos for each client. For FedVMR and federated baselines, we set the number of the local epoch as $10$ and the fraction of updated clients in each round as $100\%$. The number of communication rounds is set as $40$ for Charades-STA and $70$ for ActivityNet Captions. The average video length is about $4$ times more than that of moments. Therefore, the dataset is divided into 4 temporal classes based on the temporal position of the start timepoint, i.e., the first quarter, the second, the third, and the last. The same goes for the end timepoint. The whole dataset would be divided into 16 temporal classes by taking the position of both the start and end timepoint into consideration.


\noindent\textbf{Distribution Simulation Among Clients.} Similar to existing works \cite{fedvc,2019arxiv}, we used Dirichlet distribution to generate the non-IID data partition among clients. Note that we set the number of clients on Charades-STA as $16$ to correspond to $16$ scenes. Similarly, we set $200$ clients on ActivityNet Captions as the number of activity classes is $200$.

\begin{figure}
	\centering
	\includegraphics[width=0.43\textwidth]{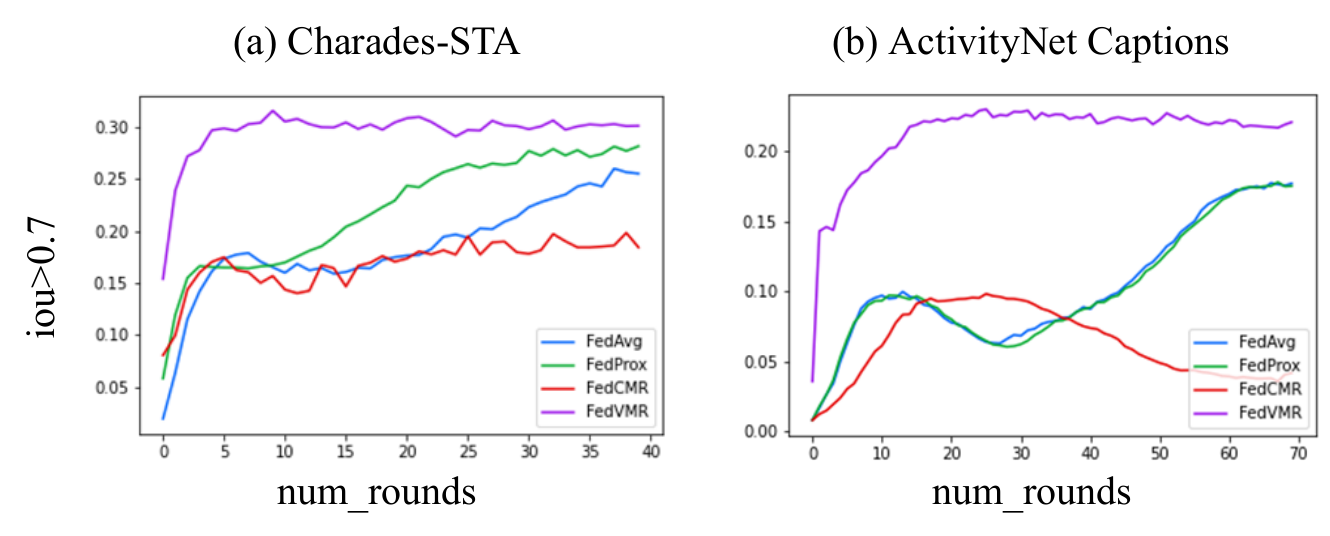}
	\caption{The test accuracy as a function of rounds.}
	\label{exp3}
\end{figure}

\begin{figure}
	\centering
	\includegraphics[width=0.25\textwidth]{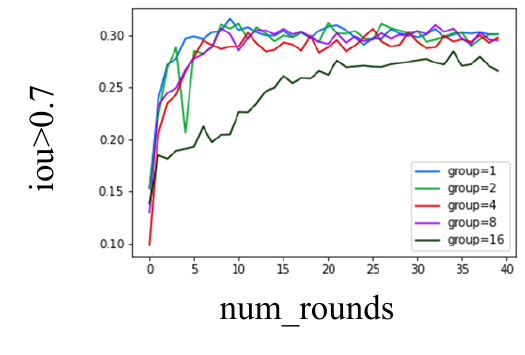}
	\caption{Experimental results of different $G$ on Charades-STA.}
	\label{exp4}
\end{figure}

\subsection{Comparison with Baselines}\label{comparison}
The performance and the convergence results are presented in Table~\ref{tab:baseline} and Figure~\ref{exp3}, respectively. The data distribution is generated with $\alpha=0$, where $\alpha$ represents the concentration parameter in the Dirichlet distribution. According to the results, we can have the following observations. (1) The performance of FedVMR can approximate or even surpass the model trained on centralized data. (2) FedVMR can reach convergence faster than baselines. (3) The curves of FedVMR are smoother than others and do not have sudden changes during the training procedure. These experimental results demonstrate that FedVMR is effective and converges fast.

\subsection{Further Analysis} \label{ablation}
We conducted ablation experiments to gain a deeper understanding of our model.

\noindent\textbf{Value of Groups.} To evaluate the influence of the different number of groups, we conducted experiments with various values of $G$, i.e., $1,2,4,8$, and $16$. The results with IoU$>0.7$ on Charades-STA are shown in Figure~\ref{exp4}. We can find $G = 1, 2, 4$, and $ 8$ visibly perform better and converge faster than the case of $G = 16$ (totally parallel), which demonstrates the effectiveness of our grouped sequential federated learning.

Similar to Figure~\ref{exp4}, we conducted experiments on Charades-STA by varying the values of groups. We recorded the rounds needed when achieving convergence and the results are listed in Table \ref{tab:efficiency}. Please note that the last line ($G = 16$) is similar to traditional federated learning, which trains all clients parallelly. By denoting the time cost of this case as $\star$, we presented the time comparisons among all cases. We can find that our grouped sequential federation learning could achieve a balance between effectiveness and efficiency.

\begin{table} 
	 \scriptsize
	\centering
	\begin{tabular}{lcr}
		\toprule[1pt]
		number of group (clients per group) & rounds needed & time comparisons\\
		\midrule
		1 (16)&    4  &    $ 1.88\times\star$ \\
		2 (8)&   3  &    $0.71\times\star$ \\
		4 (4)&    5  &    $ 0.59\times\star$ \\
		8 (2)&    6  &    $0.35\times\star$ \\
      \hdashline
		16 (1) \%traditional FL setting &   34  &   $\star$  \\
		\toprule[1pt]
	\end{tabular}
	\caption{Time comparisons with different numbers of group.}
	\label{tab:efficiency}
\end{table}


\section{Conclusion}
In this paper, we first introduce federated learning into the VMR domain. To deal with this new task, we propose a novel federated method FedVMR which contains several modules, e.g., grouped sequential learning, c-validation set, and temporal-distribution-gap loss to solve the performance degradation induced by distributed learning, local model integration and no-IIDness.  Experimental results on two benchmark datasets have demonstrated the effectiveness of our work.


\newpage
\bibliographystyle{named}
\bibliography{refs}

\end{document}